\documentclass[letterpaper]{article} 
\usepackage{aaai24}  
\usepackage{times}  
\usepackage{helvet}  
\usepackage{courier}  
\usepackage[hyphens]{url}  
\usepackage{graphicx} 
\usepackage{natbib}  
\usepackage{caption} 
\usepackage{algorithm}
\usepackage{algorithmicx}

\usepackage{style_bo}

\usepackage{newfloat}
\usepackage{listings}
\DeclareCaptionStyle{ruled}{labelfont=normalfont,labelsep=colon,strut=off} 
\floatstyle{ruled}
\newfloat{listing}{tb}{lst}{}
\floatname{listing}{Listing}
\nocopyright

\title{AutoLTS: Automating Cycling Stress Assessment via Contrastive Learning and Spatial Post-processing}
\author{
    Bo Lin$^1$,
    Shoshanna Saxe$^2$,
    Timothy C. Y. Chan$^1$
}
\affiliations{
    \textsuperscript{\rm 1}Department of Mechanical and Industrial Engineering, University of Toronto\\
    \textsuperscript{\rm 2}Department of Civil and Mineral Engineering, University of Toronto\\
    \{blin, tcychan\}@mie.utoronto.ca, s.saxe@utoronto.ca
}

\usepackage{bibentry}

\begin{document}

\maketitle

\begin{abstract}
Cycling stress assessment, which quantifies cyclists' perceived stress imposed by the built environment and motor traffics, increasingly informs cycling infrastructure planning and cycling route recommendation. However, currently calculating cycling stress is slow and data-intensive, which hinders its broader application. In this paper, We propose a deep learning framework to support accurate, fast, and large-scale cycling stress assessments for urban road networks based on street-view images. Our framework features i) a contrastive learning approach that leverages the ordinal relationship among cycling stress labels, and ii) a post-processing technique that enforces spatial smoothness into our predictions. On a dataset of 39,153 road segments collected in Toronto, Canada, our results demonstrate the effectiveness of our deep learning framework and the value of using image data for cycling stress assessment in the absence of high-quality road geometry and motor traffic data. 
\end{abstract}

\section{Introduction}

Safety and comfort concerns have been repeatedly identified as major factors that inhibit cycling uptake in cities around the world. A range of metrics, such as the level of traffic stress (LTS) \citep{furth2016network, huertas2020level} and bicycle level of service index \citep{callister2013tools}, have been proposed to quantify cyclists' perceived stress imposed by the built environment and motor traffic. These metrics are predictive of cycling behaviors \citep{imani2019cycle,wang2020commuting} and accidents \citep{chen2017bicycle}, and thus have been applied to support cycling infrastructure planning \citep{lowry2016prioritizing,gehrke2020cycling,chan2022machine} and route recommendation \citep{chen2017bicycle,castells2020cycling}. However, calculating these metrics typically requires high-resolution road network data, such as motor traffic speed, the locations of on-street parking, and the presence/type of cycling infrastructure on each road segment. The practical challenge of collecting accurate and up-to-date data hinders the broader application of cycling stress assessment and tools built on it.

\begin{figure}[!th]
    \centering
    \begin{subfigure}[b]{0.23\textwidth}
         \centering
         \includegraphics[width=0.9\textwidth]{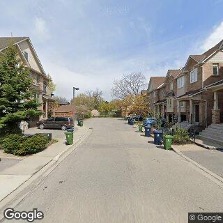}
         \caption{LTS1}
     \end{subfigure}
     \hfill
     \begin{subfigure}[b]{0.23\textwidth}
         \centering
         \includegraphics[width=0.9\textwidth]{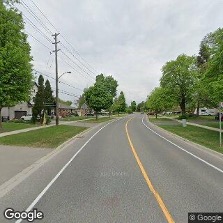}
         \caption{LTS2}
     \end{subfigure}
     \hfill
     \begin{subfigure}[b]{0.23\textwidth}
         \centering
         \includegraphics[width=0.9\textwidth]{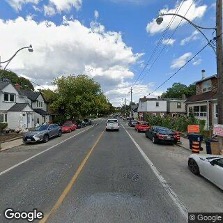}
         \caption{LTS3}
     \end{subfigure}
     \hfill
     \begin{subfigure}[b]{0.23\textwidth}
         \centering
         \includegraphics[width=0.9\textwidth]{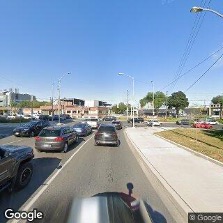}
         \caption{LTS4}
     \end{subfigure}
        \caption{Example images with the four LTS labels: LTS1 roads are safe for all cyclists including children, LTS2 roads are for most adults, LTS3 and LTS4 are for ``enthused and confident'' and ``strong and fearless'' cyclists, respectively.}
        \label{fig:lts_examples}
\end{figure}

\begin{figure*}[!ht]
    \centering
    \includegraphics[width=0.8\textwidth]{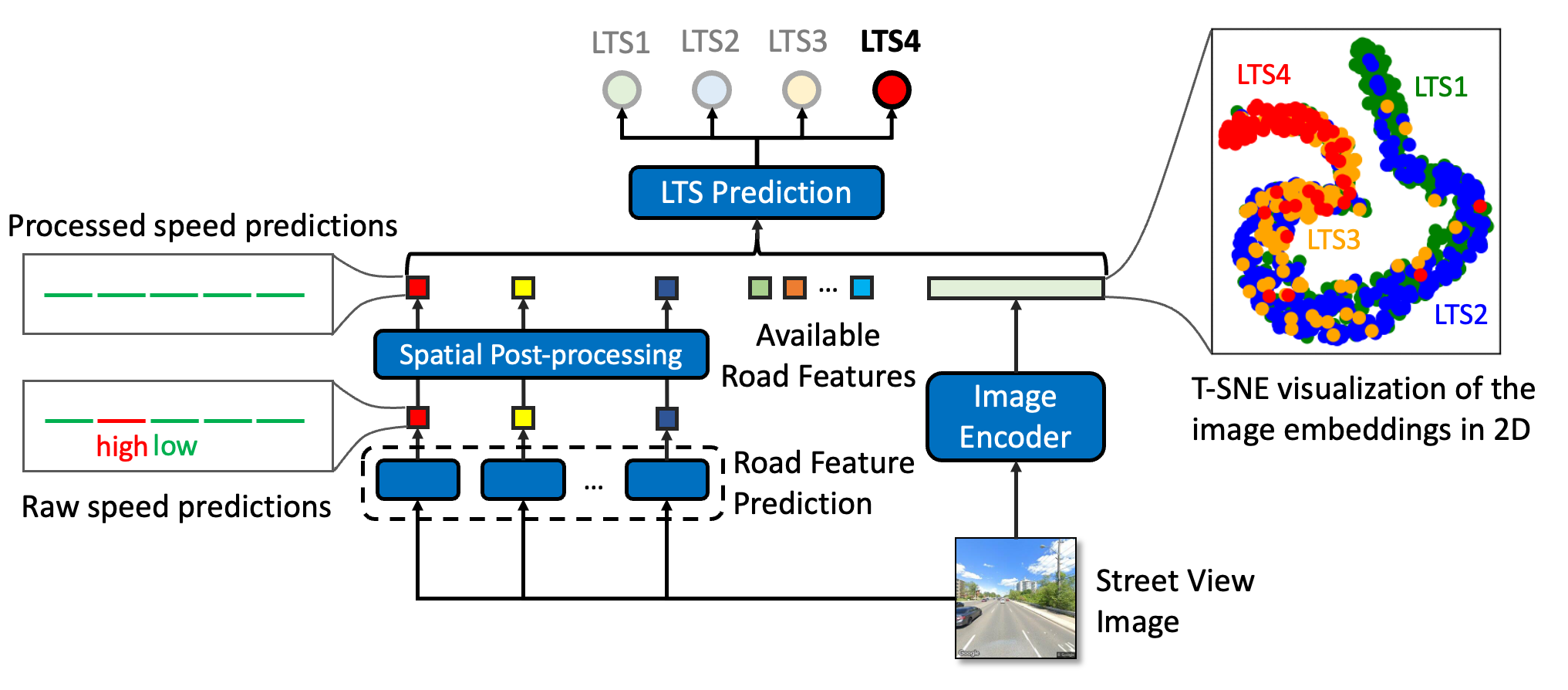}
    \caption{An overview of AutoLTS. The input image is encoded to an image embedding and is used to predict missing road features. The image encoder is trained using a contrastive learning approach (Section \ref{subsec:contrastive}). The predicted road features go through a post-processing module (Section \ref{subsec:spatial_postprocessing}) that enforces spatial smoothness into the predictions. Finally, a feedforward network predicts the the image's LTS label based on the image embedding, and the predicted and available road features.}
    \label{fig:mdl_arch}
\end{figure*}

To tackle this challenge, we propose AutoLTS, a deep learning framework for assessing cycling stress of urban road networks based on street-view images. AutoLTS can facilitate timely, accurate, and large-scale assessments of cycling stress because up-to-date street-view images are easy to access via the Google StreetView API. Using a dataset of 39,153 road segments collected in Toronto, Canada, we focus on automating the calculation of the LTS metric. Specifically, as shown in Figure \ref{fig:lts_examples}, road segments are classified into four classes, i.e., LTS 1, 2, 3 and 4 \citep{dill2016revisiting}, corresponding to the cycling suitability of four types of cyclists, where LTS 1 is the least stressful road and LTS 4 is the most stressful. This metric has been applied to investigate the connectivity \citep{lowry2016prioritizing,kent2019prioritizing} and equity \citep{tucker2018bicycle} of urban cycling networks and to evaluate cycling interventions during the COVID-19 pandemic \citep{lin2021impact}. While we focus on LTS for demonstration, our approach applies to any cycling stress metric. 

Formulating this task as a simple image classification problem may not utilize the training dataset to its full potential because it ignores i) the causal relationship between road features and LTS, ii) the ordinal relationships among LTS labels, and iii) the spatial structure of urban road networks. It is critical to leverage i)--iii) to improve the prediction performance as our dataset, limited by the practical data collection challenge and the number of road segments in a city, is relatively small for a computer vision task. Item ii) is of particular importance as misclassifications between different pairs of LTS labels carry different empirical meanings. For example, predicting an LTS1 road as LTS3 is considered worse than predicting it as LTS2 because LTS2 corresponds to the cycling stress tolerance of most adults \citep{furth2016network}. The former may lead to redundant cycling infrastructure on a low-stress road and or recommended cycling routes that exceed most adults' stress tolerance.  

As illustrated in Figure \ref{fig:mdl_arch}, to capture i), we formulate the LTS assessment as a two-step learning task. We first predict LTS related road features based on the input image and learn high-quality representations of the image. We then combine the image embedding with the predicted and available road features to produce the final LTS prediction. This two-step framework allows us to capture ii) and iii) via \textit{contrastive learning} and a \textit{spatial post-processing} technique, respectively. Specifically, to address ii), we propose a contrastive learning approach to learn an image embedding space where images are clustered based on their LTS labels, and where these clusters are positioned according to the ordinal relationship among these labels. To tackle iii), we develop a post-processing technique to enforce spatial smoothness into road feature predictions. We opt not to directly enforce spatial smoothness into LTS predictions because it may smooth over important local patterns, which are critical for downstream applications such as cycling network design that aims to fix the disconnections between low-stress sub-networks. Our contributions are summarized below.

\begin{enumerate}
    \item \textbf{A novel application.} We introduce the first dataset     and the first computer vision framework for automating cycling stress assessment.
    \item \textbf{New methodologies.} We propose a new contrastive loss for ordinal classification that generalizes the supervised contrastive loss \citep{khosla2020supervised}. We develop a post-processing technique that adjusts the road feature predictions considering the spatial structure of the road network. Both 
    can be easily generalized to other tasks.
    \item \textbf{Strong performance.} Through comprehensive experiments using a dataset collected in Toronto, Canada, we demonstrate i) the value of street-view images for cycling stress assessment, and ii) the effectiveness of our approach in a wide range of real-world settings. 
\end{enumerate}

\section{Literature Review}

\paragraph{Computer vision for predicting urban perceptions.} Street view images have been used to assess the perceived safety, wealth, and uniqueness of neighborhoods \citep{salesses2013collaborative,arietta2014city,naik2014streetscore,ordonez2014learning,dubey2016deep} and to predict neighborhood attributes such as crime rate, housing price, and voting preferences \citep{arietta2014city,gebru2017using}. We contribute to this stream of literature by i) proposing the first dataset and the deep-learning framework for assessing cycling stress, and ii) developing the first post-processing technique to enforce spatial smoothness in model predictions. Our proposal of automating cycling stress assessment via a computer vision approach is similar to the work of \cite{ito2021assessing} who use pre-trained image segmentation and object detection models to extract road features and then construct a bike-ability index based on them. In contrast, we focus on automating the calculation of a cycling stress metric that is well-validated in the transportation literature. The approach proposed by \cite{ito2021assessing} does not apply because many LTS-related road features are i) unlabeled in the dataset on which the segmentation and object detection models were trained (e.g. road and cycling infrastructure types) or ii) not observable in street-view images (e.g. motor traffic speed). 

\paragraph{Contrastive learning.} 
Contrastive learning, which learns data representations by contrasting similar and dissimilar data samples, has received growing attention in computer vision. Such techniques usually leverage a contrastive loss to guide the data encoder to pull together similar samples in an embedding space, which has been shown to facilitate downstream learning in many applications \citep{zhao2021what, bengar2021reducing, bjorck2021accelerating}, especially when data labels are unavailable or scarce. To date, most contrastive learning approaches are designed in unsupervised settings \citep{gutmann2010noise, sohn2016improved, oord2018representation,hjelm2018learning,wu2018unsupervised,bachman2019learning,he2020momentum,chen2020simple}. They typically generate ``similar'' data by applying random augmentations to unlabeled data samples. More recently, \citet{khosla2020supervised} apply contrastive learning in a supervised setting where they define ``similar'' data as data samples that share the same image label. Linear classifiers trained on the learned embeddings outperform image classifiers trained directly based on images. We extend the supervised contrastive loss \citep{khosla2020supervised} by augmenting it with terms that measure the similarity of images with ``neighboring'' labels. Consequently, the relative positions of the learned embeddings reflect the similarity between their class labels, which helps to improve our model performance.

\section{Method}

\subsection{Data Collection and Pre-processing}
Training and testing our model requires three datasets: i) road network topology, ii) ground-truth LTS labels for all road segments, and iii) street-view images that clearly present the road segments. We collect all the data in Toronto, Canada via a collaboration with the City of Toronto. Data sources and pre-processing steps are summarized below.

\textbf{Road network topology}: We retrieve the centerline road network from \citet{trt2020open}. Geospatial coordinates of both ends of each road segment are presented. We exclude roads where cycling is legally prohibited, e.g., expressways. The final network has 59,554 road segments.

\textbf{LTS label.} The LTS calculation requires detailed road network data. For each road segment in Toronto, we collect road features as summarized in Table \ref{tab:data_sources} and calculate its LTS label following \citet{furth2016network} and \citet{imani2019cycle} (detailed in Appendix \ref{app:lts_cal}).

\textbf{Street-view image.} We collect street-view images using the Google StreetView API. We opt not to collect images for road segments that are shorter than 50 meters because a significant portion of those images typically present adjacent road segments that may have different LTS labels. For each of the remaining road segments, we collect one image using the geospatial coordinate of its mid-point. We manually examine the collected images to ensure that they clearly present the associated road segments. If an image fails the human screening, we manually recollect the image when possible. Images are missing for roads where driving is prohibited, such as trails and narrow local passageways. 

\begin{table}[!ht]
\footnotesize
\centering
\begin{tabular}{@{}lll@{}}
\toprule
\multicolumn{1}{c}{Feature}  & \multicolumn{1}{c}{Source} \\ \midrule
Road type                    & \cite{trt2020open}          \\
Road direction               & \cite{trt2020open} \\
Number of lanes              & \cite{CanGov2020}           \\
Motor traffic speed                & \cite{TravelModelGroup2016}  \\
Cycling infrastructure location  & \cite{trt2020open}          \\
On-street parking location & \cite{TPA2020}      \\ \bottomrule
\end{tabular}
\caption{Summary of LTS-related road features.}
\label{tab:data_sources}

\end{table}

Our final image dataset consists of 39,153 high-quality street-view images, with 49.0\%, 34.5\%, 6.9\%, and 9.7\% of them labeled as LTS 1, 2, 3 and 4, respectively. 

\subsection{Supervised Contrastive Learning for Ordinal Classification} \label{subsec:contrastive}

\begin{figure*}[!ht]
    \centering
    \includegraphics[width=0.8\textwidth]{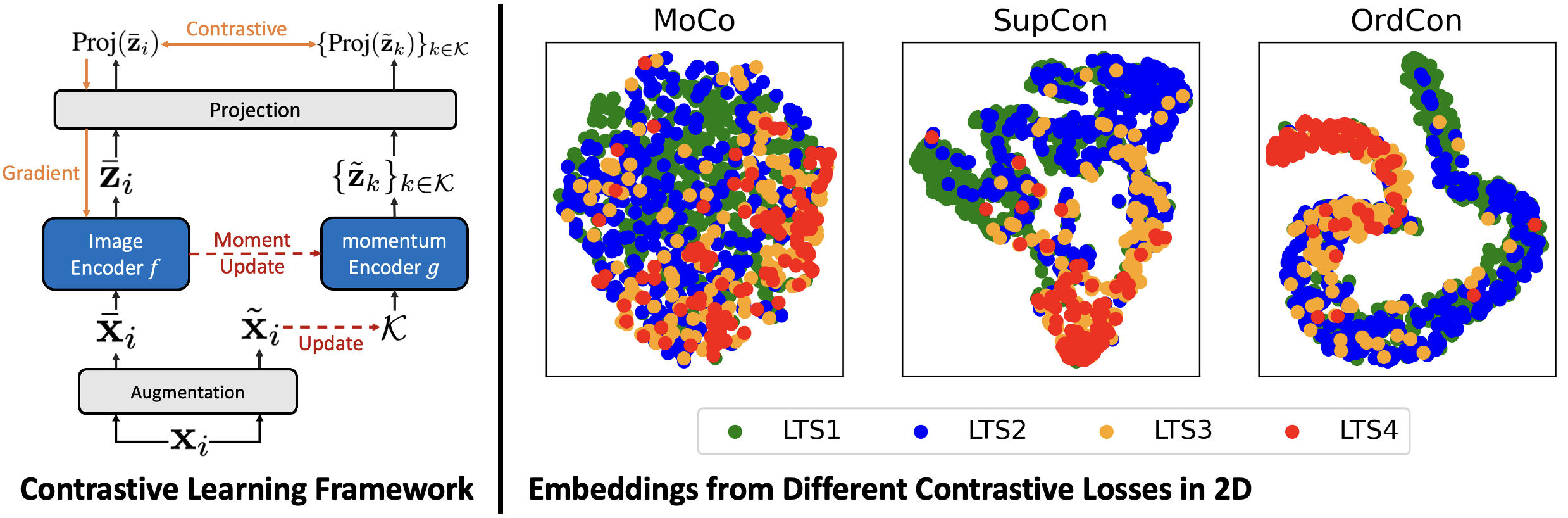}
    \caption{The contrastive learning framework and the learned image embeddings from different contrastive losses. MoCo indicates the self-supervised contrastive loss, SupCon indicates the supervised contrastive loss, and OrdCon indicates our contrastive loss. All the embeddings are projected to a two-dimensional space via T-SNE \citep{hinton2002stochastic}. Each point corresponds to one street-view image and is color-coded according to the associated LTS label.}
    \label{fig:contras_and_emb}
\end{figure*}

We propose a contrastive learning approach to train the image encoder. The novelty lies in the development of a new contrastive loss that considers the ordinal relationship among LTS labels. We adopt a contrastive learning framework (Figure \ref{fig:contras_and_emb}) similar to MoCo \citep{he2020momentum} to train the image encoder $f$ on a pretext task where the encoder learns to pull together ``similar'' images in the embedding space. Given a batch of $n$ road segments indexed by $\mN$, let $\bfx_i$ and $y_i$ denote the street view image and the label of segment $i \in \mN$, respectively. We assume $y_i\in [m]$ are discrete and ordered for all $i\in \mN$. We create $l$ virtual labels $(y_i^1, y_i^2, \ldots, y_i^l)$ for each image $\bfx_i$ where $y_i^u = \lceil y_i / u \rceil$ for all $u \in [l]$. In words, these virtual labels are created by grouping the ``neighboring'' real labels at different granularities. Consequently, images with ``similar'' real labels have more overlapping virtual labels.  We create two views $\bar{\bfx}_i$ and $\tilde{\bfx}_i$ of each image $\bfx_i$ by applying a random augmentation module twice. We create a momentum encoder $g$ that has the same structure as $f$ and whose parameters are updated using the momentum update function \citep{he2020momentum} as we train $f$. The image views $\{\bar{\bfx}_i\}_{i\in [n]}$ and $\{\tilde{\bfx}_k\}_{k\in \mK}$ are encoded by $f$ and $g$, respectively, where $\mK$ is a fixed-length queue that stores previously generated image views. Let $\bar{\bfz}_i = f(\bar{\bfx}_i)$ and $\tilde{\bfz}_i = g(\tilde{\bfx}_i)$ denote the embedding generated by these two encoders. During training, these embeddings are further fed into a projection layer, which is discarded during inference following \citet{khosla2020supervised} and \citet{chen2020simple}. The encoder network $f$ is trained to minimize the following loss that applies to the projected embeddings:

\begin{equation}
\label{loss:sup_ord}
    L^{\textrm{ord}} = 
        - \frac{1}{N} 
        \sum_{i\in \mN} 
        \sum_{u\in [l]}
        \frac{w^u}{|\mK^u_i|}
        \sum_{j\in \mK^u_i}
        \log{\frac{\exp\left[\textrm{p}(\bar{\bfz}_i)^\intercal \textrm{p}(\tilde{\bfz}_{j})/\tau\right]}
        {\sum_{k\in \mK} \exp\left[\textrm{p}(\bar{\bfz}_i)^\intercal \textrm{p}(\tilde{\bfz}_k) / \tau\right]}}.
\end{equation}
where $\mK^u_i = \{k\in \mK: y^u_i = y^u_k \}$ for all $u \in [l]$, $w^u$ is a constant weight assigned to the $u^\textrm{th}$ virtual label, $\tau$ is a temperature hyper-parameter, and $\textrm{p}$ is the projection function.

\textbf{Comparison to other loss functions}. Compared to MoCo \citep{he2020momentum}, our OrdCon takes advantage of label information. Consequently, as illustrated in Figure \ref{fig:contras_and_emb}, our image embeddings form clusters that correspond to their image labels. Compared to the SupCon \citep{khosla2020supervised}, our OrdCon considers the ordinal relationship among image labels by aggregating the real label at different granularities. As a result, the relative positions of our embedding clusters reflect the similarity between their corresponding labels. OrdCon recovers the SupCon when $l=1$ and $w^1 = 1$.

\subsection{Spatial Post-processing for Road Feature Predictions} \label{subsec:spatial_postprocessing}

Several LTS-related road features, e.g., motor traffic speed, have strong spatial correlations, meaning that the values associated with adjacent road segments are highly correlated. Such structure can be useful in regulating road feature predictions, which may lead to improved LTS predictions. However, it is often not obvious how spatial smoothness should be enforced. For example, consider a case where the motor traffic speeds of five consecutive road segments are predicted as 60, 40, 60, 40, and 60 km/h, respectively. It is likely that two of them are wrong, yet it is unclear if we should change the 40s to 60 or 60s to 40. In this section, we propose a principled way to address this problem.

\subsubsection{A Causal Model} \label{subsec:causal}

\begin{figure}[!ht]
    \centering
    \includegraphics[width=0.2\textwidth]{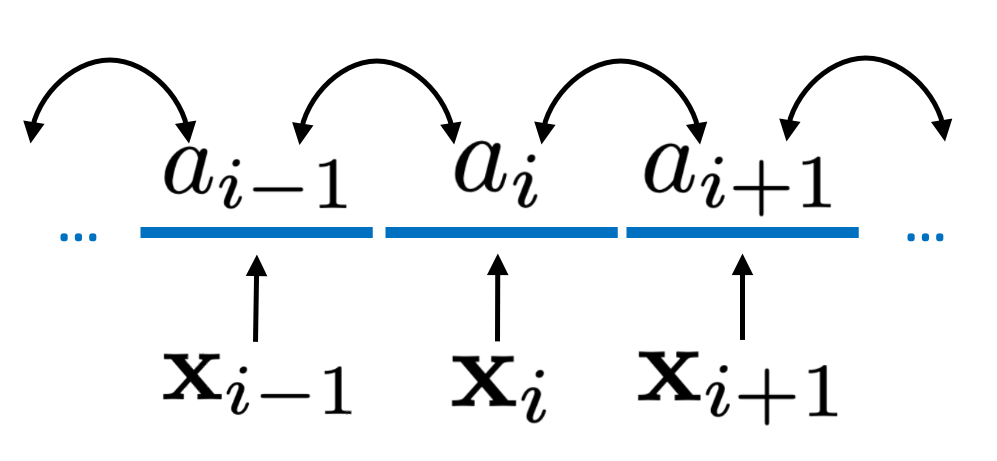}
    \caption{A causal model for road feature predictions. The blue lines indicate real-world road segments, black arrows represent causal impacts.}
    \label{fig:causal_model}
\end{figure}

We start by introducing a directed arc graph (DAG) (illustrated in Figure \ref{fig:causal_model}) that describes the relationships between the inputs $\bfx_i$ (i.e., street view images) and targets $a_i \in \mA$ (i.e., the road feature of interest) of our road-feature prediction module (illustrated in Figure \ref{fig:mdl_arch}). We assume $\mA$ to be discrete. This is not restrictive because continuous road features can be categorized according to the LTS calculation scheme (detailed in Appendix \ref{app:road_fea}). Let $\mI$ denote the set of edges in the road network and $\mJ(i) \subset \mI$ denote the set of road segments that are adjacent to road segment $i \in \mI$. We make three assumptions as listed below.

\begin{enumerate} 
    \item For any $i\in \mI$ and $k \in \mI\backslash \mJ(i)$, $a_i$ and $a_k$ are conditionally independent given $\{a_j\}_{j\in \mJ(i)}$.
    \item For any $i, j \in \mI$ and $j \neq i$, $\bfx_i$ and $a_j$ are conditionally independent given $a_i$.
    \item For any $i\in \mI$ and $j, k \in \mJ(i)$ and $j \neq k$, $a_j$ and $a_k$ are conditionally independent given $a_i$.
\end{enumerate}

The first and second assumptions state that the target $a_i$ is directly influenced only by the input of the same road segment $\bfx_i$ and the targets of its adjacent segments $\{a_j\}_{j\in \mJ(i)}$. The third assumption states that when target $a_i$ is known, its impacts on its adjacent targets $\{a_j\}_{j\in \mJ(i)}$ are independent. This model naturally applies to several LTS-related road features. For example, the traffic speed on a road segment is affected by the built environment observable from its street-view image ($\bfx_i$) and the traffic speeds on its adjacent road segments ($\{a_j\}_{j\in \mJ(i)}$). The built environment and traffic speeds on other road segments may present indirect impacts on the road segment of interest, but such impacts must transmit through its adjacent road segments. Additionally, the impacts of the traffic speed on a road segment on the speeds of its adjacent road segments can be viewed as independent (or weakly dependent) because they usually correspond to motor traffics along different directions.

\subsubsection{Enforcing Spatial Smoothness}

Given the DAG, target predictions can be jointly determined by maximizing the joint probability of all targets given all inputs, i.e., $\textrm{maximize}_{\bfa} P\left(\{a_i\}_{i\in \mI} \middle| \{\bfx_i\}_{i\in \mI} \right)$. However, evaluating the joint distribution of $\{a_i\}_{i\in \mI}$ is non-trivial because our DAG is cyclic. Instead, we look into determining the target of one road segment at a time assuming all other targets are fixed.

\begin{proposition}
    Under assumptions 1--3, for any $i\in \mI$,
    \begin{equation}
        P\left (a_i \middle| \{\bfx_i\}_{i\in \mI}, \{a_j\}_{j\neq i \in \mI} \right)
        \propto
        \prod_{j\in \mJ(i)} P\left(a_j | a_i \right) P(a_i | \bfx_i) 
    \end{equation}
    \label{prop:1}
\end{proposition}

Proposition \ref{prop:1} decomposes the conditional probability of target $a_i$ given all other targets and inputs. The transition probability $P(a_j | a_i)$ can be estimated from our training data, and $P(a_i | \bfx_i)$ can be produced by our deep learning model. The proof is presented in Appendix \ref{app:proof}. Inspired by Proposition \ref{prop:1}, we next introduce an algorithm that iteratively updates the target predictions in the whole network until there are no further changes. The algorithm is summarized in Algorithm \ref{alg:iter_update}. 

\begin{algorithm}[!ht]
\caption{An iterative target adaptation algorithm}\label{alg:iter_update}
\textbf{Input}: Initial predictions $\{a_i\}_{i\in \mI}$; Transition Probabilities $\{P(a | a')\}_{a, a' \in \mA}$; Model Predictions $\{P(a_i| \bfx_i)\}_{i\in \mI}$; Adjacent sets $\mJ(i)$ for any $i\in \mI$. \\
\textbf{Output}: Updated predictions $\{\hat{y}_i\}_{i\in \mI}$.

\begin{algorithmic}[1]
\Repeat
    \State set $\hat{a_i} \gets a_i$ for all $i \in \mI$. 
    \For{$i\in \mI$}
        \State set $a_i \gets \argmax_{a\in \mA} \prod_{j\in \mJ(i)} P\left(\hat{a}_j | a \right) P(a | \bfx_i)$
    \EndFor
\Until{{$\hat{a}_i = a_i$ for all $i \in \mI$}}
\end{algorithmic}
\end{algorithm}

\section{Empirical Results}

\subsection{Experiment Setup} \label{subsec:exp_setup}

\textbf{Evaluation scenarios.} We evaluate AutoLTS and baseline methods in three data-availability scenarios, each under four train-test-validation splits, totaling 12 sets of experiments. 

For \textit{data availability}, we consider LTS based on
\begin{enumerate}
    \item Street view image
    \item Street view image,  road and cycling infrastructure types
    \item Street view image, number of lanes, and speed limit.
\end{enumerate}
The design of these scenarios is informed by the real data collection challenges we encountered in Toronto. The number of lanes and the speed limit of each road segment are accessed via Open Data Canada \citep{CanGov2020}. Road type and the location of cycling infrastructure are available via Open Data Toronto \citep{trt2020open}. However, as the two data platforms use different base maps, combining data from these two sources requires considerable manual effort, echoing the data collection challenges in many other cities. 

\begin{figure*}[!th]
    \centering
    \includegraphics[width=.75\textwidth]{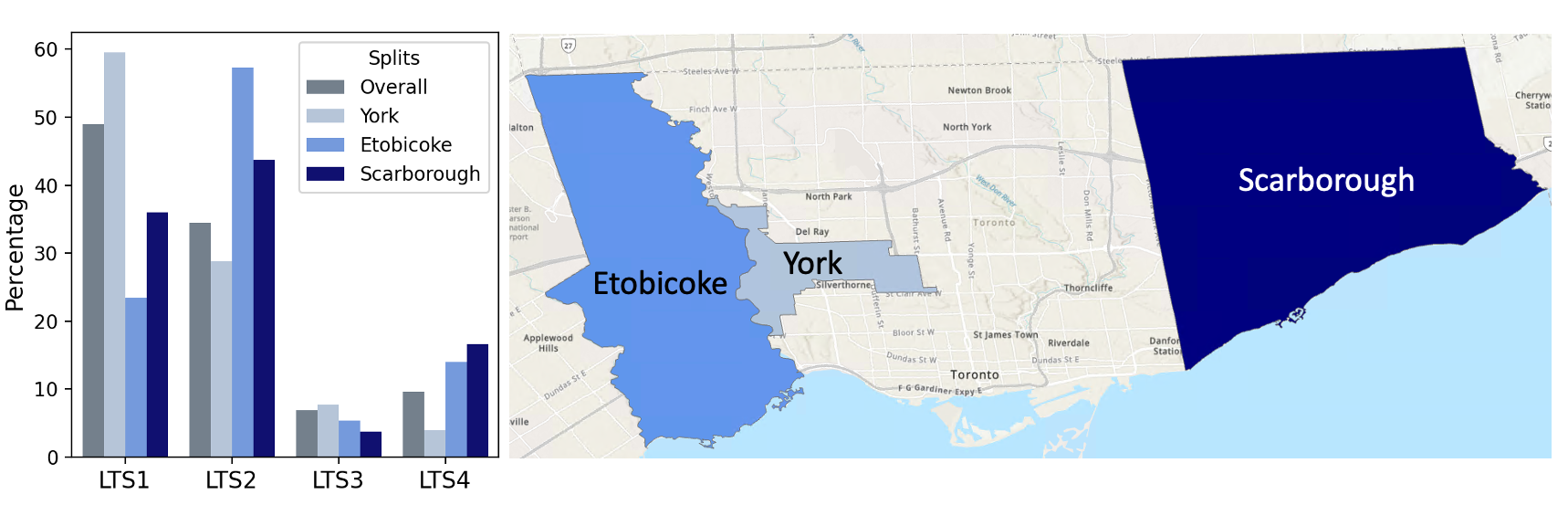}
    \caption{Illustration of the three spatial splits. York has a similar LTS distribution as the overall city-wide distribution. Etobicoke has the majority of the road segments being LTS2 and more roads being LTS4 compared to the city's average. Scarborough has an even higher LTS4 percentage.}
    \label{fig:spatial_split}
\end{figure*}

For the \textit{train-test-validation split}, we consider 
\begin{enumerate}
    \item A random 70/15/15 train-test-validation split across all road segments in Toronto.
    \item Three spatial splits, which use road segments in an area as the test set and performs a random 80/20 train-validation split for other road segments. As shown in Figure \ref{fig:spatial_split}, we consider using road segments in three of Toronto's amalgamated cities, York, Etobicoke, and Scarborough as the test sets. These three areas have very different LTS distributions, which allows us to exam the generalization ability of AutoLTS in real-world settings.
\end{enumerate}
The random split mimics the situation where we use AutoLTS to extrapolate manual LTS assessment or to update the LTS assessment in the city where the model is trained. The spatial split mimics the situation where we apply AutoLTS trained in one city to an unseen city.

\textbf{Evaluation Metrics.} 
\begin{itemize}
    \item LTS Prediction Accuracy
    \begin{equation}
        \textrm{Acc} = 
        \frac{1}{{N_\textrm{test}}}
        \sum_{i=1}^{N_\textrm{test}} 
            \mathds{1}[y_i = \hat{y}_i].
    \end{equation}
    \item High/Low-Stress Prediction Accuracy 
    \begin{equation}
        \textrm{HLA} = 
        \frac{1}{N_\textrm{test}}
        \sum_{i=1}^{N_\textrm{test}} 
            \mathds{1}[h(y_i) = h(\hat{y}_i)]
    \end{equation} where $h$ is a function that takes a value of $1$ if the input LTS label is low-stress (LTS1/2) and takes $0$ if the LTS label is high-stress (LTS3/4).
    \item Average False High/Low-Stress Rate 
    \begin{equation}
        \textrm{AFR} = 
        \frac{1}{2}
        \left\{
            \frac{\sum_{i=1}^{N_\textrm{test}}
            \mathds{1}[h(\hat{y}_i) = 1]}{n^l_{\textrm{test}}}
            + 
            \frac{\sum_{i=1}^{N_\textrm{test}}
            \mathds{1}[h(\hat{y}_i) = 0]}{n^h_{\textrm{test}}}
        \right\}
    \end{equation}
    where $n^l_{\textrm{test}}$ and $n^h_{\textrm{test}}$ denote, respectively, the numbers of test road segments that are low- and high-stress.
\end{itemize}
Acc and HLA measure the overall prediction performance, while AFR considers the fact that the dataset is imbalanced with a higher portion being low-stress. Ideally, we want a model that achieves high Acc and HLA and low AFR.

\begin{table*}[!th]
\centering
\begin{tabular}{@{}llrrrrrrrrrrrr@{}}
\toprule
&         
& \multicolumn{3}{c}{\begin{tabular}[c]{@{}c@{}}Random \\ ($N_\textrm{test}=$ 5,873)\end{tabular}} 
& \multicolumn{3}{c}{\begin{tabular}[c]{@{}c@{}}York \\ ($N_\textrm{test}=$ 2,091)\end{tabular}}    
& \multicolumn{3}{c}{\begin{tabular}[c]{@{}c@{}}Etobicoke \\ ($N_\textrm{test}=$ 6,667)\end{tabular}} 
& \multicolumn{3}{c}{\begin{tabular}[c]{@{}c@{}}Scarborough \\ ($N_\textrm{test}=$ 8,921)\end{tabular}}  
\\ \cmidrule(lr){3-5} \cmidrule(lr){6-8} \cmidrule(lr){9-11} \cmidrule(lr){12-14}
Sce.  & Method   
& Acc & HLA & AFR & Acc & HLA & AFR & Acc & HLA & AFR & Acc & HLA & AFR
\\ \midrule
\multirow{4}{*}{1} 
& Cross-Entropy  
& 70.49  & 93.51  & 10.19  
& 60.97  & 93.40  & 12.09   
& 64.20  & 92.89  & 9.37     
& 64.28  & 93.87  & 12.61       \\
& MoCo    
& 61.69  & 90.23  & 14.68      
& 57.68  & 91.34  & 17.17       
& 52.03  & 89.89  & 12.71       
& 56.16  & 91.45  & 17.01   \\
& SupCon  
& 70.75  & 93.41  & 11.73  
& 61.17  & 93.40  & 17.05       
& 64.29  & 93.19  & 9.29  
& 65.73  & 93.38  & 10.96    \\
& AutoLTS 
& \textbf{73.41} & \textbf{94.16}  & 10.50 
& \textbf{62.31} & \textbf{94.69} & 15.72 
& \textbf{64.69}  & \textbf{93.50}  & 9.87
& \textbf{66.04}  & \textbf{94.62}  & \textbf{10.77} \\ 
\midrule
\multirow{5}{*}{2} 
& CART    
& 56.21  & 96.87  & 5.21
& 43.33  & 96.75  & 4.51
& 35.35  & 96.73  & 7.25       
& 50.22  & 96.40  & 8.34           \\
& Cross-Entropy  
& 75.07  & 96.82  & 5.91   
& 63.37  & 96.37  & 5.26  
& 66.21  & 95.97  & 6.55       
& 67.76  & 95.74  & 10.14       \\
& MoCo    
& 68.94  & 96.65  & 14.41
& 57.39  & 96.22  & 4.81
& 59.11  & 95.79  & 9.08       
& 62.09  & 96.31  & 8.66        \\
& SupCon  
& 74.89  & 96.42  & 6.33 
& 64.13  & 96.17  & 5.54
& 65.70  & 95.68  & 9.14       
& 68.55  & 96.19  & 8.66        \\
& AutoLTS 
& \textbf{75.86} & 96.22  & 7.02
& \textbf{65.04} & 96.13  & 8.77 
& \textbf{67.74} & 96.20  & 7.37
& \textbf{68.86}  & \textbf{96.51}  & \textbf{7.78}     \\ \midrule
\multirow{5}{*}{3} 
& CART    
& 89.41 & 96.07   & 10.08       
& 88.81 & 97.57   & 6.97       
& 90.67 & 95.46   & 10.37   
& 91.90 & 94.90   & 12.51 \\
& Cross-Entropy  
& 90.26  & 95.33  & 8.97        
& 88.12  & 97.37  & 6.47       
& 91.01  & 95.34  & 7.79
& 91.45  & 95.54  & 12.88 \\
& MoCo    
& 89.82  & 95.37  & 11.25       
& 86.90  & 98.61  & 3.63       
& 90.88  & 96.35  & 6.78  
& 92.74  & 94.91  & 12.79    \\
& SupCon  
& 91.20  & 96.19  & 11.30   
& 87.42  & 96.70  & 4.18      
& 89.04  & 95.93  & 7.69 
& 92.41  & 95.11  & 11.50 \\
& AutoLTS 
& \textbf{91.65} & \textbf{96.70}  & \textbf{5.87}      
& \textbf{89.24}  & 97.23  & 4.78   
& \textbf{92.61}  & \textbf{96.68}  & \textbf{4.77} 
& \textbf{94.50}  & \textbf{97.28}  & \textbf{5.81} \\ \bottomrule
\end{tabular}
\caption{The out-of-sample performance of AutoLTS and baselines. The three blocks (top to bottom) correspond to data-availability scenarios 1, 2, and 3, respectively (Section \ref{subsec:exp_setup}). The four groups of columns (left to right) correspond to the train-test-validation splits defined in Section \ref{subsec:exp_setup}. Numbers in boldface are cases where our approach achieves the best performance.} \label{tab:main_res}
\end{table*}

\textbf{Baselines.} To demonstrate the value of image data, in scenarios where road features are available, we use a classification and regression tree (CART) that predicts LTS based on available road features as a baseline. CART is selected because the LTS calculation scheme \citep{furth2016network} can be summarized by a decision tree. We also compare AutoLTS with image-based supervised and contrastive learning methods. For supervised learning, we consider Res-50 \citep{he2016deep} trained using the cross-entropy loss. For contrastive learning, we consider supervised contrastive learning (MoCo) \citep{he2020momentum} and self-supervised contrastive learning (SupCon) \citep{khosla2020supervised}, both implemented with the MoCo trick \citep{he2020momentum}. Baselines are detailed in Appendix \ref{app:baseline}.

\textbf{Model details.} We use ResNet-50 \citep{he2016deep} as the image encoder. The normalized ReLU activations of the final pooling layer are used as the image embedding ($\xi$=2,048). We follow \citet{he2020momentum} to set $\tau = 0.07$ and use the SimCLR augmentation \citep{chen2020simple} for training. We train one ResNet-50 to predict each missing road feature. All road features are discretized using the thresholds defined in the LTS calculation scheme \citep{furth2016network} (Appendix \ref{app:road_fea}). In the LTS prediction module, we first train a CART model to predict a road segment's LTS based on its predicted and available road features. We then use the LTS distribution in the leaf node that a road segment is assigned to as its road feature embedding, which is mapped to a $\xi$-dimensional space via a linear layer and averaged with the image embedding. Finally, a linear classifier predicts the road segment's LTS based on the averaged embedding. Training details are summarized in Appendix \ref{app:autoLTS_training}.

\begin{table*}[!th]
\centering
\begin{tabular}{@{}lrrrrrrrrrrrr@{}}
\toprule  
& \multicolumn{3}{c}{\begin{tabular}[c]{@{}c@{}}Random \\ ($N_\textrm{test}=$ 5,873)\end{tabular}} 
& \multicolumn{3}{c}{\begin{tabular}[c]{@{}c@{}}York \\ ($N_\textrm{test}=$ 2,091)\end{tabular}}   
& \multicolumn{3}{c}{\begin{tabular}[c]{@{}c@{}}Etobicoke \\ ($N_\textrm{test}=$ 6,667)\end{tabular}} 
& \multicolumn{3}{c}{\begin{tabular}[c]{@{}c@{}}Scarborough \\ ($N_\textrm{test}=$ 8,921)\end{tabular}}  
\\ \cmidrule(lr){2-4} \cmidrule(lr){5-7} \cmidrule(lr){8-10} \cmidrule(lr){11-13} 
Model & Acc & HLA & AFR & Acc & HLA & AFR & Acc & HLA & AFR & Acc & HLA & AFR
\\ \midrule
2Step-Exact  
& 41.97  & 52.41  & 33.89      
& 57.05  & 84.41  & 29.92         
& 23.23  & 42.31  & 39.09
& 24.18  & 35.90  & 45.31  \\ 
2Step-Spatial-Exact  
& 43.10  & 54.20  & 32.28       
& 58.06  & 85.41  & 21.60            
& 23.46  & 44.10  & 38.81
& 25.39  & 37.79  & 43.12 \\   \midrule
MoCo-NN   
& 61.69  & 90.23  & 14.68 
& 57.68  & 91.34  & 17.17         
& 52.03  & 89.89  & 12.71       
& 56.16  & 91.45  & 17.01  \\ 
SupCon-NN 
& 70.75  & 93.41  & 11.73     
& 61.17  & 93.40  & 17.05             
& 64.29  & 93.59  & 9.45 
& 65.73  & 93.38  & 10.96 \\ 
OrdCon-NN
& 71.11  & 93.96  & 9.95      
& 60.74  & 93.93  & 11.62 
& 64.02  & 93.98  & 9.29 
& 65.95  & 94.54  & 10.55 \\  \midrule
AutoLTS-MoCo 
& 72.21  & 93.70  & 10.11      
& 61.60  & 94.02  & 13.52            
& 64.69  & 92.94  & 9.84  
& 64.40  & 94.05  & 11.62  \\ 
AutoLTS-SupCon
& 73.30  & 94.16  & 10.61      
& 62.17  & 94.38  & 15.76  
& 64.63  & 93.42  & 10.66  
& 65.86  & 94.37  & 11.43  \\ 
AutoLTS-OrdCon
& 73.41 & 94.16  & 10.50 
& 62.31 & 94.69  & 15.72 
& 64.69  & 93.50 & 9.87
& 66.04  & 94.62 & 10.77   \\ 
\bottomrule
\end{tabular}
\caption{Summary of ablation studies.} \label{tab:ablation}
\end{table*}

\subsection{Main Results}

The performance of AutoLTS and baselines are shown in Table \ref{tab:main_res}. We summarize our findings below.

\textbf{The value of image data for cycling stress assessment}. AutoLTS achieves LTS prediction accuracy of 62.31\%--73.41\% and high/low-stress accuracy of 93.50\%--94.69\% only using street-view images. Such a model can be useful for cycling infrastructure planning and route recommendation tools that do not require the granularity of four LTS categories and focus solely on the difference between high- (LTS3/4) and low-stress (LTS1/2) road segments. In data-availability scenarios where partial road features are available (scenarios two and three), incorporating street-view images leads to increases of 0.43--32.39 percentage points in Acc with little to no increases in AFR. The improvements are particularly large in scenario two where the average increase in Acc due to the usage of street-view images is 23.10 percentage points across all train-test-validation splits considered. By combining street-view images with the speed limit and the number of lanes (scenario 3), the Acc is over 90\% under all splits. These numbers demonstrate that street view images are valuable for cycling stress assessment with and without partial road features.

\textbf{The performance of AutoLTS and other image-based methods.} 
Overall, AutoLTS achieves the highest Acc, which is of primary interest, in all evaluation scenarios. Due to the limited sample size, unsupervised contrastive learning (MoCo) generally falls around 10\% behind SupCon. SupCon outperforms the simple image classification formulation (Cross-Entropy) in 8 out of 12 scenarios yet is inferior to AutoLTS in all scenarios. However, we observe that when there is a significant domain shift from training to test data (spatial splits), all methods including AutoLTS are more prone to overfitting the training data, and thus have worse out-of-sample performance than in random splits.  

\subsection{Ablation Studies}

Next, we present ablation studies using data-availability scenario one to demonstrate the values of our two-step learning framework, ordinal contrastive learning loss, and the post-processing module, The results are summarized in Table \ref{tab:ablation}.

\textbf{The value of the two-step learning framework.} We compare AutoLTS with an alternative approach that replaces the LTS prediction module with the exact LTS calculation scheme (2Step-Exact and 2Step-Spatial-Exact). This change leads to reductions of 6.16--40.83 percentage points in Acc due to the compounded errors from the first step, highlighting the importance of second-step learning. Moreover, AutoLTS outperforms all baselines that predict LTS based only on image (MoCo-, SupCon-, and OrdCon-NN), demonstrating the value of incorporating road feature predictions. 

\textbf{The value of ordinal contrastive learning.} 
We compare the three contrastive learning methods using the AutoLTS framework and the linear classification protocol \citep{he2020momentum}. When used to predict LTS without road features (MoCo-, SupCon-, and OrdCon-NN), OrdCon and SupCon are competitive in Acc, yet OrdCon constantly achieves higher HLA and lower AFR because it considers the relationship among LTS labels. This is practically important because the ability to distinguish between low- and high-stress roads plays a vital role in most adults' cycling decision makings \citep{furth2016network}. When combined with road features, all contrastive learning methods perform reasonably well. Nevertheless, Auto-OrdCon consistently outperforms others by a meaningful margin.

\textbf{The value of spatial post-processing.} Applying the spatial post-processing technique to road feature predictions generally leads to an increase of around 1\% in road feature prediction accuracy (presented in Appendix \ref{app:road_fea}) which can be translated into improvements in LTS prediction Acc (2Step-Exact versus 2Step-Spatial-Exact). While the improvement seems to be limited, it corresponds to correctly assessing the LTS of 21--162 road segments in the studied area, which can have a significant impact on the routing and cycling infrastructure planning decisions derived based on the assessment.

\section{Conclusion}

In this paper, we present a deep learning framework, AutoLTS, that uses streetview images to automate cycling stress assessment. AutoLTS features i) a contrastive learning approach that learns image representations that preserve the ordinal relationship among image labels and ii) a post-processing technique that enforces spatial smoothness into the predictions. We show that AutoLTS can assist in accurate, timely, and large-scale cycling stress assessment in the absence of road network data. 

Our paper has three limitations, underscoring potential future research directions. First, we observe performance degradation when the training and test data have very different label distributions (spatial splits). Future research may apply domain adaptation methods to boost the performance of AutoLTS in such scenarios. Second, AutoLTS does not consider the specific needs of downstream applications. For instance, in cycling route recommendations, under-estimation of cycling stress may be more harmful than over-estimation because the former may lead to cycling routes that exceed cyclists' stress tolerance and result in increased risks of cycling accidents. In cycling network design, cycling stress predictions might be more important on major roads than on side streets because cycling infrastructure is typically constructed on major roads. Such impacts may be captured by modifying the loss function to incorporate decision errors. Finally, all our experiments are based on a dataset collected in Toronto. Future research may collect a more comprehensive dataset to further assess the generalizability of our model. We hope this work will open the door to using deep learning to support the broader application of cycling stress assessment and to inform real-world decision makings that improve transportation safety and efficiency.


\bibliography{paper}

\clearpage
\appendix

\section{Proof of Statements} \label{app:proof}

\subsection{Proof of Proposition \ref{prop:1}}
\begin{proof}
We have

\begin{align*}
    P\left
        (a_i  
        \middle|
        \bfx_i, \{a_j\}_{j\neq i \in \mI}
    \right)
    & = \frac{
    P\left(
        \{a_j\}_{j\in \mJ(i)} 
        \middle|
        \bfx_i, a_i 
    \right)
    P(\bfx_i, a_i)
    }
    {P\left(
        \{a_j\}_{j\in \mJ(i)} 
        \middle| 
        \bfx_i
    \right)
    P(\bfx_i)
    } \\
    & = \frac{
    \prod_{j\in \mJ(i)} P\left(
        a_j \middle| a_i 
    \right)
    P(a_i | \bfx_i) 
    P(\bfx_i)
    }
    {P\left(
        \{a_j\}_{j\in \mJ(i)} 
        \middle|
        \bfx_i
    \right)
    P(\bfx_i)
    } \\
    & = \frac{
    \prod_{j\in \mJ(i)} P\left(
        a_j | a_i 
    \right)
    P(a_i | \bfx_i) 
    }
    {P\left(
        \{a_j\}_{j\in \mJ(i)} 
        \middle| 
        \bfx_i
    \right)
    }.
\end{align*}

The first equation follows the definition of conditional probability. The second equation holds because of the assumptions 1, 2, and 3 presented in Section \ref{subsec:causal}. Since the denominator in  the last line is a constant, we have 
\begin{equation}
    P\left (a_i \middle| \bfx_i, \{a_j\}_{j\in \mJ(i)} \right)
    \propto
    \prod_{j\in \mJ(i)} P\left(a_j | a_i \right) P(a_i | \bfx_i) 
\end{equation}
\end{proof}

\section{LTS Calculation Details} \label{app:lts_cal}
We follow \citet{furth2016network} and \citet{imani2019cycle} to calculate the LTS label of every road segment in Toronto. The calculation scheme can be summarized by the following decision rules, which are applied in sequence.

\begin{itemize}
    \item Road segments that are multi-use pathways, walkway, or trails are LTS 1.
    \item Road segment with cycle tracks (i.e. protected bike lanes) are LTS 1.
    \item For road segments with painted bike lanes:
    \begin{itemize}
        \item If the road segment has on-street parking, 
        \begin{itemize}
            \item If one lane per direction and motor traffic speed $\leq$ 40 km/h, then LTS 1.
            \item If one lane per direction and motor traffic speed $\leq$ 48 km/h, then LTS 2.
            \item If motor traffic speed $\leq$ 56 km/h, then LTS 3.
            \item Otherwise, LTS4.
        \end{itemize}
        \item If the road segment has no on-street parking, 
        \begin{itemize}
            \item If one lane per direction and motor traffic speed $\leq$ 48 km/h, then LTS 1.
            \item If one/two lanes per direction, then LTS 2.
            \item If motor traffic speed $\leq$ 56 km/h, then LTS 3
            \item Otherwise, LTS 4.
        \end{itemize}
    \end{itemize}
    \item For road segments without cycling infrastructure:
    \begin{itemize}
        \item If motor traffic speed $\leq$ 40 km/h, and $\leq$ 3 lanes in both directions, 
        \begin{itemize}
            \item If daily motor traffc volume $\leq$ 3000, then LTS 1.
            \item Otherwise, LTS 2.
        \end{itemize}
        \item If motor traffic speed $\leq$ 48 km/h, and $\leq$ 3 lanes in both directions, 
        \begin{itemize}
            \item If daily motor traffc volume $\leq$ 3000, then LTS 2.
            \item Otherwise, LTS 3.
        \end{itemize} 
        \item If motor traffic speed $\leq$ 40 km/h, and $\leq$ 5 lanes in both directions, then LTS 3.
        \item Otherwise, LTS 4.
    \end{itemize}
\end{itemize}

\section{Road Feature Prediction Details} \label{app:road_fea}

\subsection{Label Discretization}

We discretize all the road features as summarized in Table \ref{tab:discretization}. Threshold values and feature categories are selected following \citet{furth2016network} and \citet{imani2019cycle}. All road feature prediction problems are then formulated as image classfication problems.

\begin{table*}[!ht]
\centering
\caption{Road feature discretization.} \label{tab:discretization}
\begin{tabular}{@{}lcl@{}}
\toprule
\multicolumn{1}{c}{Road Feature}  & Label & \multicolumn{1}{c}{Definition}  \\ \midrule
\multirow{4}{*}{Motor traffic speed} 
& 1   & $\leq$ 40 km/h  \\
& 2   & 40 -- 48 km/h   \\
& 3   & 48 -- 56 km/h   \\
& 4   & $\geq 56$ km/h  \\ 
\midrule
\multirow{3}{*}{Road type}           
& 1   & Major/minor arterial, arterial ramp   \\ 
& 2   & Collector, access road, laneway, local road, others \\ 
& 3   & Trail, walkway        \\ 
\midrule
\multirow{5}{*}{Number of lanes} 
& 1   & One lane in both directions   \\ 
& 2   & Two lanes in both directions \\ 
& 3   & Three lanes in both directions         \\ 
& 4   & Four lanes in both directions    \\ 
& 5   & More than 5 lanes in both directions    \\ 
\midrule
\multirow{2}{*}{Road direction} 
& 1   & Unidirectional road  \\ 
& 2   & Bidirectional road \\ 
\midrule
\multirow{4}{*}{Cycling infrastructure type} 
& 1   & Bike Lane  \\ 
& 2   & Cycle track \\
& 3   & Multi-use pathway \\
& 4   & Others or no cycling infrastructure \\ 
\midrule
\multirow{2}{*}{On-street parking} 
& 1   & Has on-street parking  \\ 
& 2   & No on-street parking \\
\bottomrule
\end{tabular}
\end{table*}

\subsection{Model Details}
We train one ResNet-50 \citep{he2016deep} to predict each road feature based on the input streetview image. We initialize the model with the weights pre-trained on the ImageNet. We replace the final layer with a fully connected layer whose size corresponds to the number of possible discrete labels for the road feature. We train the model with a standard cross-entropy loss. 

\subsection{Prediction Performance}

We first present the road feature prediction accuracy under the random train-test-validation split (Tabel \ref{tab:fea_pred_random}). We compare the performance of Res50 with a naive approach that predicts road features as the corresponding majority classes observed in the training set. We observe that Res50 provides improvements of 1.86\%-22.31\% for all road features except road direction. This is because the road direction labels are highly imbalanced with over 94\% of the road segments being bi-directional. The Res50 model is able to identify some uni-directional road segments at the cost of miss-predicting some bi-directional road segments as uni-directional. We opt to use Res50 despite that it has a lower prediction accuracy than the naive approach because it leads to better prediction performance for AutoLTS according to our experiments.

\begin{table}[!ht]
\centering
\caption{Road feature prediction accuracy (\%) on the random test set ($N_\textrm{test} =$ 5,873). The ``Diff'' column highlights the improvement from ``Naive'' to ``Res50''.} \label{tab:fea_pred_random}
\begin{tabular}{@{}lrrr@{}}
\toprule
Road Feature           & Res50       & Naive   & Diff.    \\ \midrule
Road type              & 89.97    & 67.66   & +22.31   \\
Motor traffic speed    & 71.28    & 54.48   & +16.80   \\
Motor traffic volume   & 88.13    & 70.81   & +17.32   \\
Number of lanes        & 82.66    & 66.34   & +16.32   \\
Cycling infrastructure & 95.30    & 94.35   & +0.95   \\
On-street parking      & 96.05    & 94.19   & +1.86   \\
Road direction                & 94.04    & 94.18   & -0.14   \\ \bottomrule
\end{tabular}
\end{table}

We next present the road feature prediction results for all train-test-validation splits considered (Table \ref{tab:fea_pred_all}). The model performance is similar across different splits, with minor changes in prediction accuracy due to the changes in label distributions.

\begin{table*}[!ht]
\centering
\caption{Road feature prediction accuracy (\%) under all train-test-validation splits.} \label{tab:fea_pred_all}
\begin{tabular}{@{}lrrrr@{}}
\toprule
Road Feature     & Random   & York    & Etobicoke & Scarborough    \\ \midrule
Road type        & 89.97    & 90.34   & 89.47     & 88.72   \\
Motor traffic speed            & 70.19    & 65.61   & 59.52     & 57.35   \\
Number of lanes          & 82.66    & 75.32   & 85.05     & 88.05   \\
Cycling infrastructure   & 95.30    & 96.46   & 94.59     & 95.90   \\
One-street parking          & 96.05    & 95.36   & 98.16     & 99.74   \\
Road direction           & 94.04    & 85.75   & 96.10     & 99.41  \\ \bottomrule
\end{tabular}
\end{table*}

\begin{table*}[!ht]
\centering
\caption{The Prediction performance of OrdCon-NN in data-availability scenario one under the random train-test-validation split with different values of $(w^1, w^2)$.} \label{tab:impact_of_weights}
\begin{tabular}{@{}crrrrr@{}}
\toprule
$(w^1, w^2)$ & (1.00, 0.00) & (0.95, 0.05) & (0.90, 0.10) & (0.85, 0.15) & (0.80, 0.20) \\ \midrule
Acc & 70.75 & \textbf{71.11} & 69.92 & 70.78 & 70.25    \\
HLA & 93.41 & \textbf{93.96} & 93.50 & 93.51 & 93.73    \\
AFR & 11.73 & \textbf{9.95}  & 10.72 & 10.93 & 10.68   \\\bottomrule
\end{tabular}
\end{table*}

We apply the spatial post-processing module to the traffic speed prediction. As presented in Table \ref{tab:spatial}, applying the spatial post-processing technique leads to improvements of 1.01--2.05 percentage points in traffic speed prediction accuracy, corresponding to 21--162 road segments.

\begin{table*}[!ht]
\centering
\caption{Traffic speed prediction accuracy before and after spatial post-processing.} \label{tab:spatial}
\begin{tabular}{@{}lllrr@{}}
\toprule
Feature & Sub-network & Original  & Spatial\\ \midrule  
\multirow{7}{*}{\begin{tabular}[c]{@{}l@{}} Traffic \\ Speed\end{tabular}}
& \begin{tabular}[c]{@{}l@{}} Random\\ ($n_\textrm{test} = 5,873$)\end{tabular}
& \begin{tabular}[c]{@{}r@{}} 70.19\%\\ (4,122)\end{tabular}     
& \begin{tabular}[c]{@{}r@{}}$+1.29\%$\\ ($+78$)\end{tabular}  
\\
& \begin{tabular}[c]{@{}l@{}}York\\ ($n_\textrm{test} = 2,091$)\end{tabular}
& \begin{tabular}[c]{@{}r@{}} 65.61\%\\ (1,372)\end{tabular}     
& \begin{tabular}[c]{@{}r@{}}$+1.01\%$\\ ($+21$)\end{tabular}  
\\ 
&\begin{tabular}[c]{@{}l@{}}Etobicoke \\ ($n_\textrm{test} = 6,667$)\end{tabular}
& \begin{tabular}[c]{@{}r@{}}59.52\%\\ (3,968)\end{tabular}         
& \begin{tabular}[c]{@{}r@{}}$+2.05$\%\\ ($+137$)\end{tabular} 
\\
&\begin{tabular}[c]{@{}l@{}}Scarborough \\ ($n_\textrm{test} = 8,921$)\end{tabular}
& \begin{tabular}[c]{@{}r@{}}57.47\%\\ (5,127)\end{tabular}                 
& \begin{tabular}[c]{@{}r@{}}$+1.82$\%\\ ($+162$)\end{tabular}  
\\ 
\bottomrule
\end{tabular}
\end{table*}

\section{AutoLTS Training Details} \label{app:autoLTS_training}

The image encoder is trained using an SGD optimizer with an initial learning rate of 30, a weight decay of 0.0001, and a mini-batch size of 256 on an A40 GPU with RAM of 24 GB. The road feature prediction models and the LTS prediction model are trained with an SGD optimizer with an initial learning rate of 0.0003, a weight decay of 0.0001, and a mini-batch size of 128 on a P100 GPU with RAM of 12 GB. These hyper-parameters are chosen based on random search. The model is trained for 100 epochs, which takes roughly 17 hours.

We set $l$ to 2 because, according to the original LTS calculation scheme \citep{furth2016network}, the four LTS labels can be grouped into low-stress (LTS1 and LTS2) and high-stress (LTS3 and LTS4). We search for $(w^1, w^2)$ in $\{(1, 0),  (0.95, 0.05), (0.90, 0.10), (0.85, 0.15), (0.80, 0.20)\}$ and evaluate using the linear classification protocol \citet{he2020momentum} on the validation set. For example, Table \ref{tab:impact_of_weights} presents the performance of OrdCon-NN under different choices of $(w^1, w^2)$. We observe that OrdCon always helps to improve HLA and AFR, which is unsurprising because it has an additional term in the loss function to contrast low-stress and high-stress images. OrdCon also helps to enhance Acc when $(w^1, w^2)$ is set to $(0.95, 0.05)$. We use $(w^1, w^2) = (0.95, 0.05)$ for all scenarios. Further fine tuning is possible but is beyond the scope of this work due to the computational cost.

\section{Baseline Details} \label{app:baseline}

\subsection{Supervised Learning}

\subsubsection{CART}
In data-availability scenarios 2 and 3, we train a CART model to predict the LTS of a road segment based on its available road features. Hyper-parameters are selected using a grid search strategy and evaluated using a 10-fold cross-validation procedure. We summarize the hyper-parameters and their candidate values below. 
\begin{itemize}
    \item Splitting criterion: entropy, gini
    \item Max depth: 1, 2, 3, 4, 5, 6, 7, 8, 9, 10
    \item Minimum sample split: 0.01, 0.03, 0.05, 0.1, 0.15, 0.2, 2, 4, 6
\end{itemize}

\subsubsection{Res50}

\begin{figure}[!ht]
    \centering
    \includegraphics[width=0.5\textwidth]{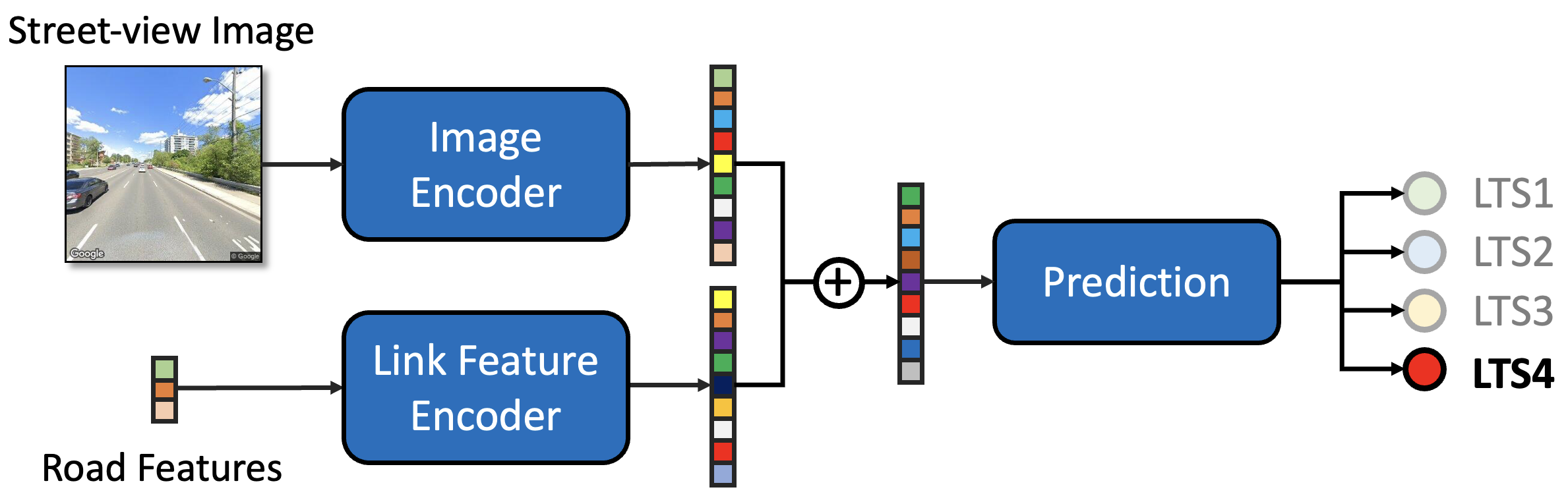}
    \caption{Model architecture.}
    \label{fig:res50_sup}
\end{figure}

We adapt the ResNet-50 model \citep{he2016deep} to predict the LTS of a road link based on its street-view image and link features that are available. As illustrated in Figure \ref{fig:res50_sup}, our model consists of three modules: 

\begin{itemize}
    \item \textbf{Image encoder.} This module extracts useful information from the street view image and represents it as a 64-dimensional vector. We implement this module with a ResNet-50 encoder followed by two fully connected layers of sizes 128 and 64, respectively.
    \item \textbf{Link-feature encoder.} This module allows us to incorporate link features when they are available. Performing link feature embedding prevents the prediction module from being dominated by the image embedding, which is higher dimensional compared to the original link feature vector. We implement this module with a fully connected layer whose input size depends on the dimensionality of the feature vector and the output size equals 64.
    \item \textbf{Prediction.} This module takes as inputs the average of the image embedding and the link-feature embedding and outputs a four-dimensional vector representing the probability of the link being classified as LTS1--4, respectively. We implement this module with a fully connected layer whose input and out sizes are set to 64 and 4, respectively.
\end{itemize}

All fully connected layers are implemented with the ReLU activation. This model is trained with an SGD optimizer with an initial learning rate of 0.0001, a weight decay of 0.0001, and a mini-batch size of 128 on a P100 GPU. Hyper-parameters are chosen using random search.

\subsection{Contrastive Learning}

\subsubsection{MoCo}
We train the image encoder $f$ depicted in Figure \ref{fig:contras_and_emb} to minimize the following loss function:

\begin{equation}
\label{loss:unsup}
    L = 
        - \frac{1}{N} 
        \sum_{i\in \mN} 
        \frac{1}{|\mK|}
        \log{\frac{\exp\left[\textrm{proj}(\bar{\bfz}_i)^\intercal \textrm{proj}(\tilde{\bfz}_{i'})/\tau\right]}
        {\sum_{k\in \mK} \exp\left[\textrm{proj}(\bar{\bfz}_i)^\intercal \textrm{proj}(\tilde{\bfz}_k) / \tau\right]}}
\end{equation}
where $i'$ is the index of the image view in $\mK$ that corresponds to the same original image as view $i$. We follow \citet{he2020momentum} to set $\tau=0.07$. We set the queue length to 25,600 and the mini-batch size to 256, which are the maximum size that can be fed into an A40 GPU. The image encoder is trained for 100 epochs, which takes roughly 34 hours. Unlike SupCon and OrdCon, MoCo is trained on all data (without a train-test-validation split) because it does not utilize the label information.

\subsubsection{SupCon}

We train the image encoder $f$ depicted in Figure \ref{fig:contras_and_emb} to minimize the following loss function:

\begin{equation}
    L = 
        - \frac{1}{N} 
        \sum_{i\in \mN}
        \frac{1}{|\mK_i|}
        \sum_{j\in \mK_i}
        \log{\frac{\exp\left[\textrm{proj}(\bar{\bfz}_i)^\intercal \textrm{proj}(\tilde{\bfz}_{j})/\tau\right]}
        {\sum_{k\in \mK} \exp\left[\textrm{proj}(\bar{\bfz}_i)^\intercal \textrm{proj}(\tilde{\bfz}_k) / \tau\right]}}
\end{equation}
where $\mK_i = \{k\in \mK: y_k = y_i\}$. We adopt the same hyper-parameters as we used for MoCo. We train the model for 100 epochs, which also takes roughly 34 hours for each evaluation scenario. We then select the model that achieves the lowest validation loss for final evaluation.

\end{document}